\documentclass[letterpaper, 10 pt, conference]{ieeeconf}  

\IEEEoverridecommandlockouts                              

\usepackage{epsfig} 

\usepackage[colorlinks]{hyperref}
\hypersetup{
    linkcolor=red,
    citecolor=green,
    urlcolor=magenta,
}
\makeatletter
\let\NAT@parse\undefined
\makeatother
\usepackage[numbers]{natbib}

\usepackage{multirow}
\usepackage{booktabs} 

\title{\LARGE \bf
Exercise Motion Classification from Large-Scale Wearable Sensor Data Using Convolutional Neural Networks
}

\author{Terry Taewoong Um$^{1}$, Vahid Babakeshizadeh$^{2}$ and Dana Kuli\'{c}$^{1}$ 
\thanks{$^{1}$Terry Taewoong Um and Dana Kuli\'{c} are with the Department of Electrical \& Computer Engineering, University of Waterloo, Waterloo, ON, N2L 3G1, Canada.
        (email: {\tt\small terry.t.um@gmail.com; dana.kulic@uwaterloo.ca})}
\thanks{$^{2}$Vahid Babakeshizadeh is with Push Inc., Toronto, ON, M5B 2G9, Canada.
        (email: {\tt\small vahid@trainwithpush.com})}
}

\begin{document}

\maketitle
\thispagestyle{empty}
\pagestyle{empty}

\begin{abstract}

The ability to accurately identify human activities is essential for developing automatic rehabilitation and sports training systems. In this paper, large-scale exercise motion data obtained from a forearm-worn wearable sensor are classified with a convolutional neural network (CNN). Time-series data consisting of accelerometer and orientation measurements are formatted as \emph{images}, allowing the CNN to automatically extract discriminative features. A comparative study on the effects of image formatting and different CNN architectures is also presented. The best performing configuration classifies 50 gym exercises with 92.1\% accuracy.
\end{abstract}

\section{INTRODUCTION}

In recent years, several computer \cite{HCICoach} and robotic \cite{RobotCoach1, RobotCoach2} systems have been proposed to coach users during rehabilitation or physical training. The coaching systems demonstrate exercises to the users through videos or robot motions and evaluate the users' movements to determine if they are performing the exercises correctly. However, the ability of the current coaching systems to recognize human movements is limited to a small number of distinctive movements. To enable these coaching systems to be used for a wider array of exercise regimens in both sports training and rehabilitation, it is necessary for the systems to be equipped with better recognition algorithms to classify a large number of exercises.

Classification of human motions, called human activity recognition (HAR), has been widely investigated in the past decade, based on vision data \cite{ActVision}, 3D data \cite{Act3D}, wearable sensor data \cite{ActWear1}, etc. Wearable sensors provide a convenient way to collect data without the need for extensive installation and setup, thus, they are often preferred when data should be collected in real life settings. Moreover, the rapid growth of the wearable device market provides access to a large amount of wearable sensor data. In this research, we take  advantage of large-scale wearable sensor data to enhance classification ability.

In large-scale data classification problems, convolutional neural networks (CNN) have demonstrated excellent performance over the past decade (e.g. \cite{AlexNet}). Although CNNs were originally proposed for static 2D images \cite{LeNet1998}, they have been successfully extended to other nonstatic domains, e.g., speech recognition \cite{CNN_Speech} and sentence classification in natural language \cite{CNN_Sentence}. In these works, CNNs learn useful features directly from raw data, i.e., no hand-crafted features are needed. As these researches have shown that learned features from CNNs can outperform classical hand-crafted features in various domains, we may expect that CNNs can also learn useful features from raw wearable sensor data and use these features for successful classification, if large-scale data is available.

\begin{figure}[!t]
	\centering
    \includegraphics[trim={0cm 1cm 0cm, 0cm},clip,scale=0.3]{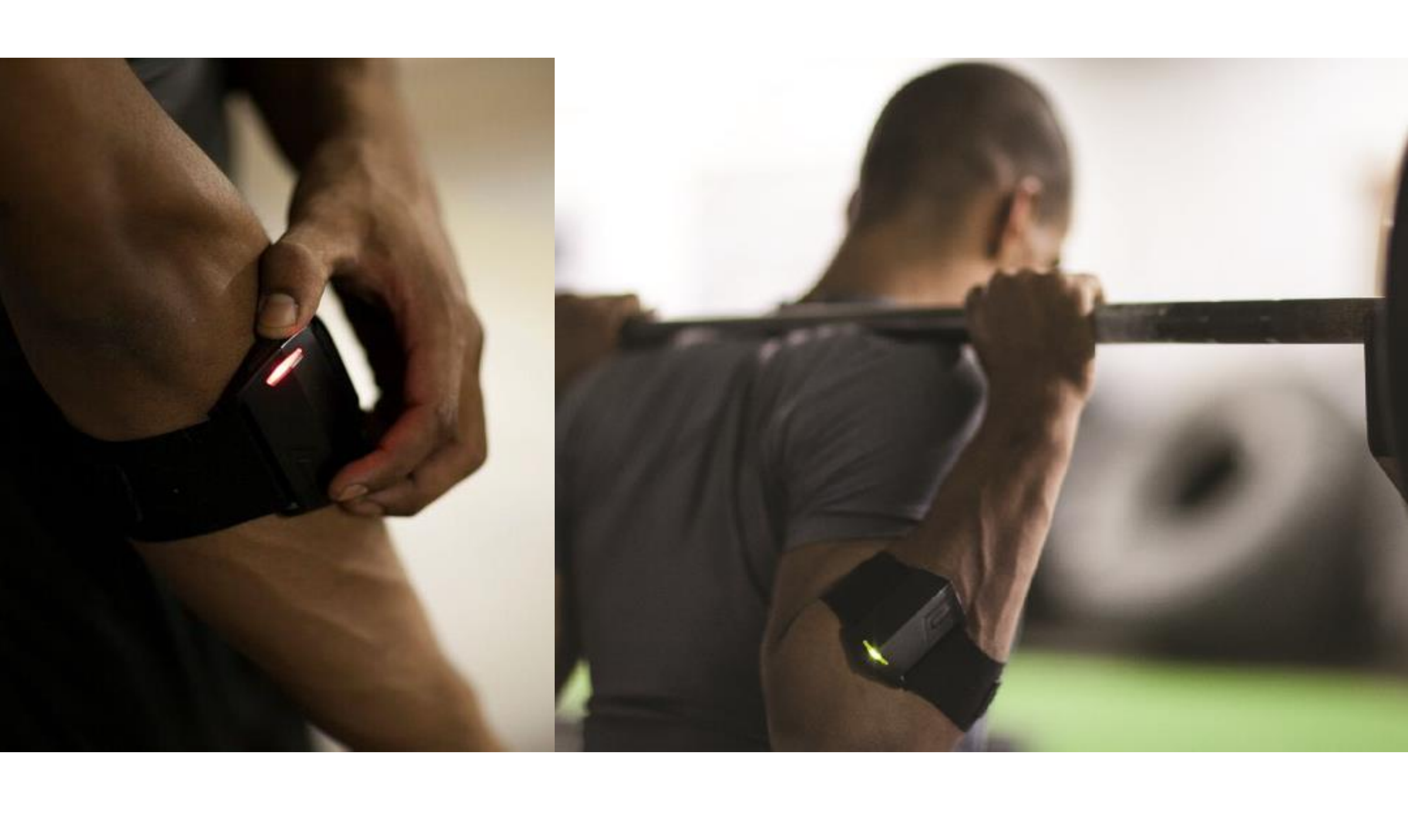}
    \caption{The wearable band, PUSH \cite{PUSH}, is a device for measuring performance of gym exercises with an accelerometer and gyroscope.}
    \label{fig:PUSH}
\end{figure}

In this research, exercise motion data obtained from a forearm band, \emph{PUSH} \cite{PUSH} (Fig. \ref{fig:PUSH}), are classified using CNN. The exercise motion data, collected from gym exercises performed by athletes, are considerably more difficult to classify than usual daily activity data, since many gym exercises have similar arm movements (e.g. front and back lat pull-down). Moreover, there are a large number of unique exercise classes (more than 300), which is a significantly larger number of classes than usual HAR with a single wearable sensor \cite{ActWear1}.

The challenging large-scale classification problem is tackled by using CNN in this research. We employed CNNs rather than recurrent neural networks (RNNs) \cite{RNN_1989}, which are usually employed for sequence data, because exercise repetition (rep) data are relatively short (mostly less than 4 seconds), and therefore, may not need long-term memory for training. By creating image-like 2D data from the raw acceleration and orientation data, CNN efficiently learns discriminative features and a classifier for differentiating between the exercises. The proposed approach can be used for computer or robotic exercise coaching systems as well as for human trainers to assist with training management and performance monitoring.

The outline of the paper is as follows: First, related research on classification with single arm-worn wearable sensor data are briefly surveyed in Section \ref{sec_works}. In Section \ref{sec_CNN}, the CNN architecture used for classifying wearable sensor data is described, including a description of the dataset. Classification results with various CNN architectures and input formats are presented in Section \ref{sec_experiment}. Finally, findings from the research and directions for future work are summarized in Section \ref{sec_conclusion}.

\section{Related work} \label{sec_works}

Although most research on HAR with wearable sensors uses multiple sensors attached to different body parts, there have been several works based on a single arm-worn sensor. In \cite{Work2009}, 5 activities (walking, running, cycling, driving, and sports) are classified with 72.3\% accuracy using decision tree, naive Bayes and naive Bayes with PCA. In this work, 19 features are first extracted from time, frequency and spatial domains and classified with the aforementioned classifiers. In \cite{Work2011}, 5 activities (running, walking, lying, standing and sitting) are classified with 91.1\% accuracy by using 13 statistical features from time and frequency domains and a multilayer perceptron classifier.

Recent works have attempted to classify more than 5 activities based on a single arm-worn sensor. \cite{Work2013} classifies 9 daily activities (brushing teeth, washing dishes, watching TV, etc.) with 82.8\% accuracy by using 12 statistical features from an accelerometer and a support vector machine classifier. They also increase the accuracy to 90.2\% by using an additional 2 features from a temperature sensor and altimeter. Lastly, \cite{Work2015} shows that 26 activities including ambulation, cycling, sedentary poses, and others, can be classified with 84.7\% accuracy using 13 statistical features and a support vector machine classifier.

The most relevant previous work to our research is \emph{RecoFit} \cite{RecoFit}, a wearable sensor to recognize and count repetitive exercises. Similar to \emph{PUSH}, \emph{RecoFit} measures 3-axis accelerometer and gyroscope data from the upper forearm. The collected data are first smoothed with a Butterworth filter and 60 statistical features are extracted from each of 5-second sliding windows. The extracted features for each window are classified using a multiclass support vector machine, and finally, the prediction is made by voting with the prediction results from all windows.

The \emph{RecoFit} system gives 100\%, 99.3\%, 98.1\%, 96.0\% accuracy for 4, 4, 7, 13-class classification, respectively. It is an impressive result, however, there are several points to remark for comparison with our research. First, they collected data from a controlled experiment environment while we used real-world data obtained from actual training sessions of athletes. Although each exercise session used in \cite{RecoFit} lasts at least 20 seconds, our data have many sets of exercises shorter than 20 seconds, and 11.5\% of the data contains fewer than four repetitions. As presented in Section \ref{sec_experiment}, few-rep data are more difficult to classify than many-rep data.

Furthermore, our research classifies 50 classes that contain many similar exercises, while the \emph{RecoFit} research classifies at most 13 classes, which are relatively distinctive (See Appendix). For example, 7 kinds of bench press and 4 kinds of squat, which could be confusing even to humans, are considered as distinct labels in our experiment. On the other hand, the 13 exercises in the \emph{RecoFit} research are relatively well-spread over a range of body-part exercises. Hence, the exercise motion classification in this research can be regarded as a more challenging and realistic problem.

\section{CNN for wearable sensor data} \label{sec_CNN}

\begin{figure}[!t]
	\centering
    \includegraphics[trim={0.4cm, 8cm, 2cm, 0cm},clip,scale=0.32]{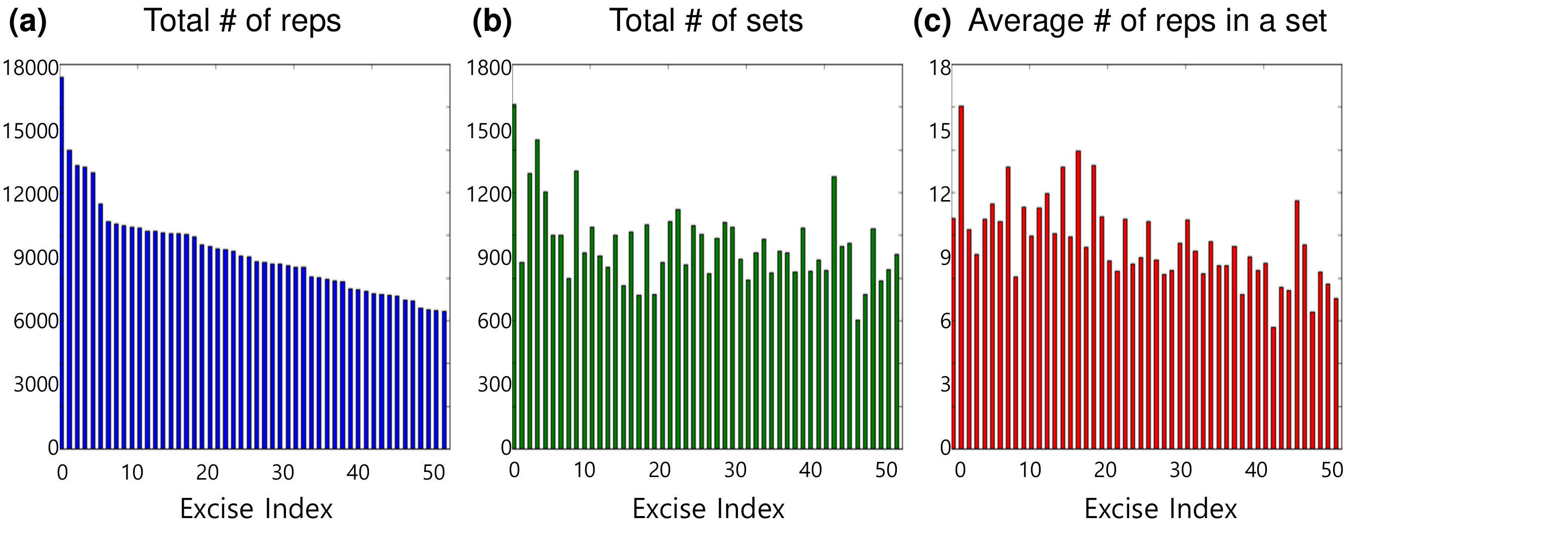}
    \caption{\emph{PUSH} data consist of 449,260 reps from 49,194 sets of exercises. The above graphs show (a) the total number of reps,  (b) the number of sets and (c) the average number of reps in a set, for each exercise. The index of the top-50 exercises are sorted by the number of reps as in (a).}
    \label{fig:NumData}
\end{figure}

\subsection{PUSH dataset} \label{sec_dataset}
\emph{PUSH} \cite{PUSH} is a forearm-worn wearable device for measuring athletes' exercise motions. \emph{PUSH} is actively being used by over fifty professional sports teams; athletes have been voluntarily collecting exercise motion data for enhancing their training performance. From the privately available dataset provided by \emph{PUSH Inc.}, we used a subset of the data consisting of the top 50 most frequently-performed exercises out of more than 300 exercises for this research (See Appendix).

The top-50-exercise dataset consists of 49,194 sets and 449,260 reps of exercises obtained from 1441 male and 307 female athletes (Fig. \ref{fig:NumData}). Each set consists of one or more reps of a single exercise, self-labeled by the athlete. Note that segmentation is performed by the device in preprocessing to extract each rep. Training and prediction process are performed based on each rep data. The number of sets and reps are necessarily unbalanced for each exercise; the exercise that has the most reps is \texttt{0:Standing triceps extension with dumbbell} (1613 sets, 17380 reps) while the exercise that has the fewest reps is \texttt{49:Wide-grip bench press with barbell} (772 sets, 6133 reps). The number of reps in a set is also necessarily varied, from 1 to 254 reps, although 94\% of the data have under 20 reps in a set. The lengths of the reps are also varied, but 99\% are shorter than 784 samples, which is equivalent to 3.92 seconds.

Since each rep has a different length, while CNN takes fixed-size data, each rep is zero-padded, i.e., filled by zeros at the tail to make the length 784 samples. The reps that have more than 784 samples are simply cropped to have 784 samples. Zero-padding is often used for solving the variable-length problem in time-series data (e.g. \cite{CNN_Sentence}). Since zero-padding preserves the length information, it is helpful for discriminating between short duration and long duration exercises.

The device is tightly bound on the upper forearm and measures accelerations and orientations with a built-in accelerometer and gyroscope, respectively. As a result, time-series data with 9 features - \texttt{(Acc\_x, Acc\_y, Acc\_z)} in the local frame, \texttt{(Acc\_x, Acc\_y, Acc\_z)} in the world frame and \texttt{EulerAngle\_x, EulerAngle\_y, EulerAngle\_z)} in the world frame - are obtained with a 200 Hz sampling rate.

\subsection{CNN Architecture} \label{sec_CNNArchi}

\begin{figure}[!t]
	\centering
    \includegraphics[trim={0cm 0.5cm 0cm, 0cm},clip,scale=0.4]{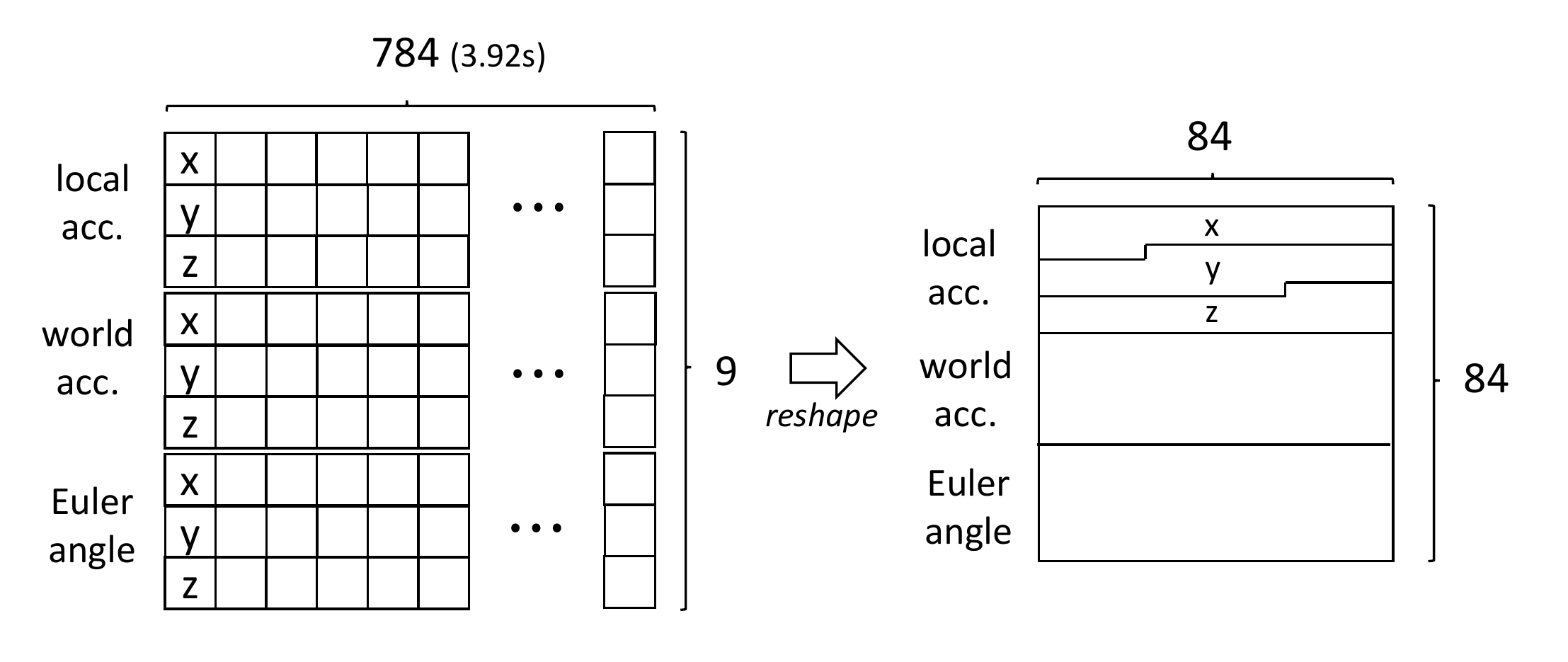}
    \caption{A 84 by 84 square image is created by reshaping the 9 by 784 raw data.}
    \label{fig:DataShape}
\end{figure}

\begin{figure}[!t]
	\centering
    \includegraphics[scale=0.40]{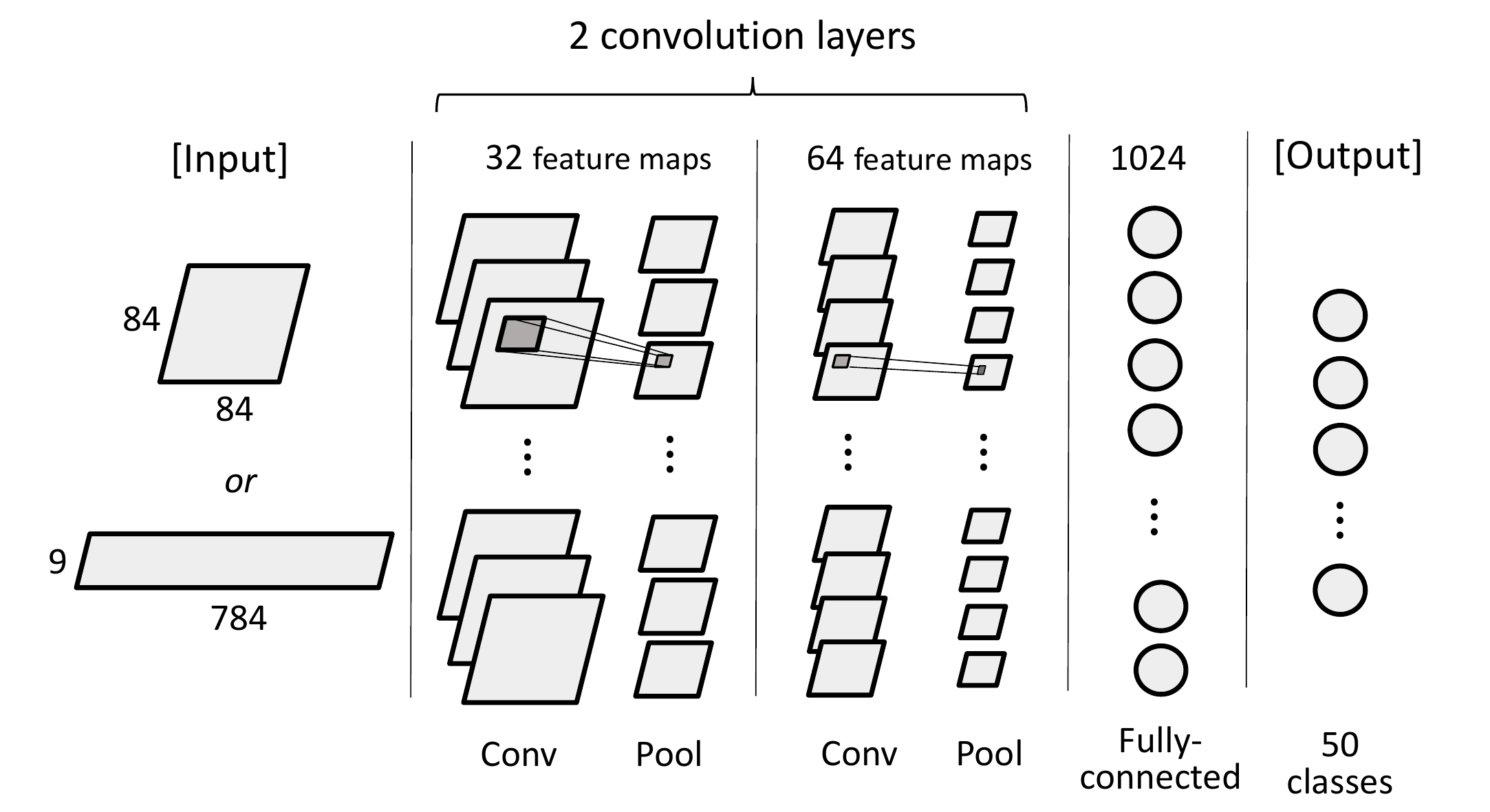}
    \caption{The baseline CNN architecture has two convolutional layers followed by a fully connected layer. In the experiments, additional convolutional layers and fully connected layers are examined.}
    \label{fig:NNArchitecture}
\end{figure}

CNN was originally developed for 2D image recognition \cite{LeNet1989}. Rather than using hand-crafted features, CNN directly learns features from raw pixels without any prior knowledge about features. CNN sweeps convolution and pooling windows over the image to create various feature maps. The convolution window convolves pixel values in a local region, called a receptive field, to determine the  spatial correlation between them. After that, the pooling window downsamples the convolved data by, e.g., selecting maximum values only. 

To apply CNN to the human motion data, we first need to create 2D \emph{images} from the raw sensor data. We create 2D images with three different approaches: (1) regard the 9 by 784 time-series data as a rectangular 2D image, (2) treat the three different feature groups - local accelerations, world accelerations, and Euler angles - as RGB channels in an image and create a 3 by 784 by 3 tensor, (3) reshape the 9 by 784 time-series data into an 84 by 84 square matrix (Fig. \ref{fig:DataShape}). Note that different choices of image formatting lead to different convolutions with different neighboring elements, which may include convolutions between irrelevant elements. 

There are several hyperparameters to be chosen for the CNN architecture; depth and width of the CNN architecture, convolution and pooling window sizes, their stride sizes, activation functions, etc. The baseline CNN architecture that is used for the experiments is presented in Fig. \ref{fig:NNArchitecture}. In the baseline CNN architecture, two convolution layers, which have 32 and 64 feature maps, are followed by a fully connected layer which has 1024 nodes. Rectified units \cite{relu} are employed as activation functions and softmax functions are used for evaluating the final 50 output node values. Experiments with different CNN architectures and input formatting will be presented in the next section.

For optimization, the Adam optimizer \cite{Adam} is employed with a learning rate of $0.0005$. Also, dropout \cite{Dropout} with a probability of $0.5$ is applied to each layer to avoid excessive dependency on certain nodes. Cross-entropy loss is employed for learning one-hot-encoded 50 exercises. Although the CNN is learned based on rep-based predictions, we also provide prediction results for each set by taking a majority vote of the rep-based predictions in each set. It is expected that set-based predictions give better results than rep-based predictions because they include the benefit of majority voting.

\begin{figure}[!t]
	\centering
    \includegraphics[scale=0.5]{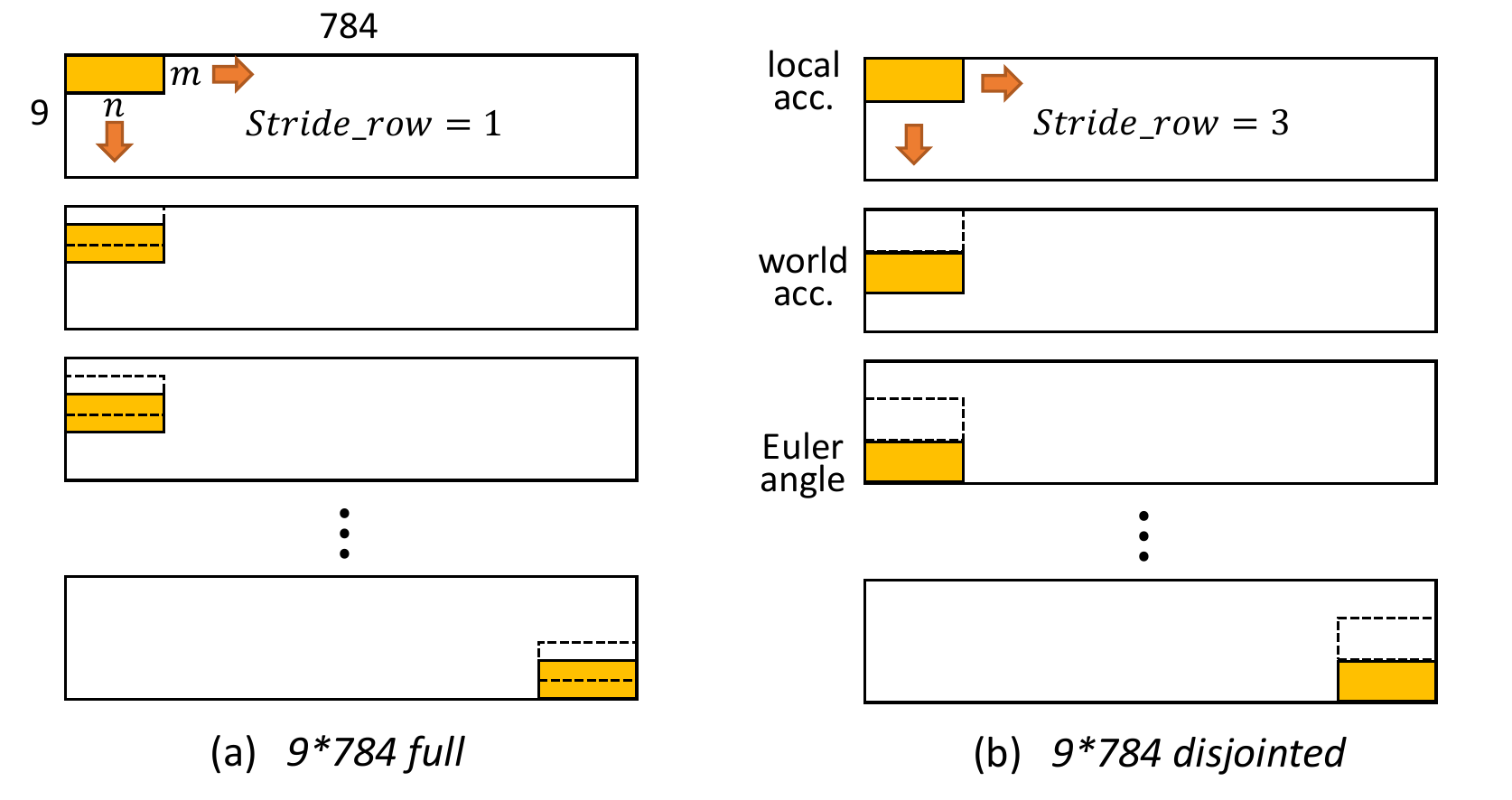}
    \caption{If the stride size along the y-axis is smaller than three as in (a), the windows will convolve elements from different feature groups, e.g., \texttt{(Acc\_y, Acc\_z, EulerAngle\_x)} . On the other hand, if the stride size along the y-axis is three as in (b), the windows will convolve elements within the same feature group.}

    \label{fig:Overlap}
\end{figure}

\section{Experiments} \label{sec_experiment}

\begin{table*}[!t]
\centering
\caption{Effects of various image formattings and convolutions: It appears that treating different feature groups as different channel (3*3*784) shows the best performance.}
\label{Table1}
\begin{tabular}{@{}cccccccccccccc@{}}
\toprule
- & Input  & \multicolumn{3}{c}{84*84} & \multicolumn{3}{c}{9*784 full}  & \multicolumn{3}{c}{9*784 disj}  & \multicolumn{3}{c}{3*784*3}
\\ \midrule \multirow{7}{*}{\begin{tabular}[c]{@{}c@{}}CNN\\ Archi-\\ tecture\end{tabular}} & -      & Win  & Stride    & Dim    & Win  & Stride & Dim
& Win  & Stride    & Dim  & Win  & Stride    & Dim \\ \cmidrule(l){2-14}
& conv1  & 3*3  & {[}1,1{]} & 84*84  & 3*3     & {[}1,1{]} & 9*784   & 3*3   & {[}3,1{]} & 3*784  & 3*3     & {[}1,1{]} & 3*784*3  \\
& pool1  & 3*3  & {[}3,3{]} & 28*28  & 3*4     & {[}3,4{]} & 3*196   & 1*4   & {[}1,4{]} & 3*196  & 2*4     & {[}1,4{]} & 2*196*3  \\
& conv2  & 3*3  & {[}1,1{]} & 28*28  & 3*3     & {[}1,1{]} & 3*196   & 3*3   & {[}1,1{]} & 3*196  & 3*3     & {[}1,1{]} & 2*196*3  \\
& pool2  & 2*2  & {[}2,2{]} & 14*14  & 1*4     & {[}1,4{]} & 3*49    & 1*4   & {[}1,4{]} & 3*49   & 2*4     & {[}1,4{]} & 1*49*3   \\ \midrule
 &        & Rep & Set      & \begin{tabular}[c]{@{}c@{}}Set\\ (\textgreater7reps)\end{tabular} & Rep & Set  & \begin{tabular}[c]{@{}c@{}}Set\\
(\textgreater7reps)\end{tabular} & Rep & Set & \begin{tabular}[c]{@{}c@{}}Set\\ (\textgreater7reps)\end{tabular} & Rep & Set  & \begin{tabular}[c]{@{}c@{}}Set\\ (\textgreater7reps)\end{tabular} \\ \midrule \multirow{2}{*}{\begin{tabular}[c]{@{}c@{}}Accuracy\\ (\%)\end{tabular}}
& Train  & 99.54   & 99.90     & 99.96  &   99.58   &  99.89   & 99.99 & 99.54 & 99.88  & 99.96   &  98.64  & 99.53    & 99.75   \\
& Test   & 81.46   & 84.86     & 89.01  &   81.20   & 85.47   &  88.85 & 81.70  & 85.61  & 89.39  & \textbf{83.34}  & \textbf{87.04} & \textbf{90.47}  \\ \midrule
\end{tabular}
\end{table*}

\begin{figure*}
	\centering
    \includegraphics[trim={0cm 10.5cm 0cm, 0cm},clip,scale=0.5]{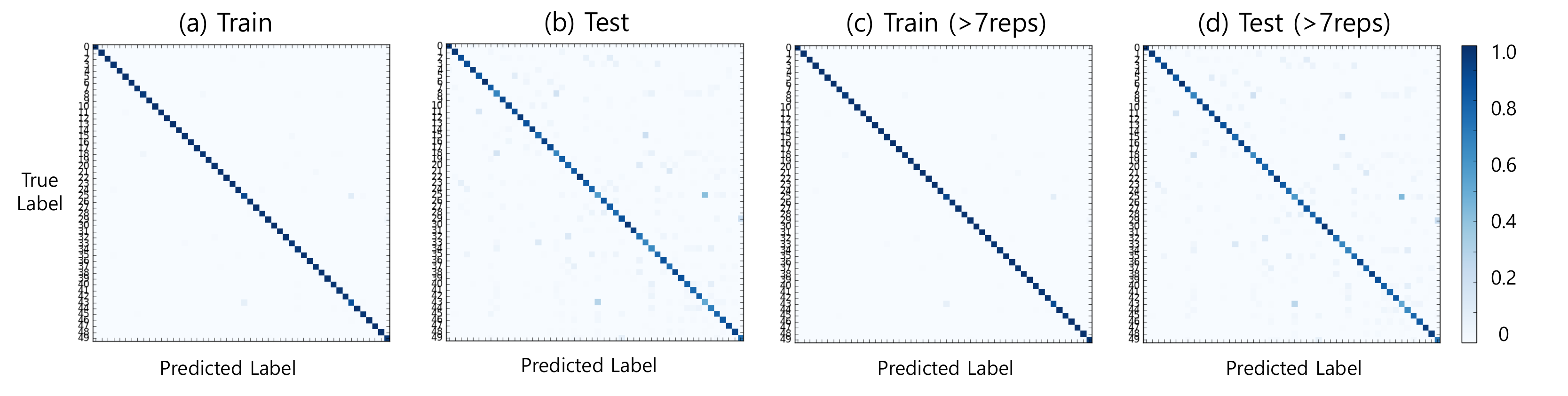}
    \caption{Normalized confusion matrices for the 3*3*784 experiment. High classification accuracy is achieved for the majority of exercises, as seen by the dominant diagonal. \texttt{43:Declined bench press with barbell} exercises are often misclassified as \texttt{25:Alternating LEG ROMANIAN deadlift with dumbbell} exercises.}
    \label{fig:Confusion}
\end{figure*}

Th experiments are performed with an open source deep learning library, Tensorflow \cite{TF}. Built-in Adam optimizer and dropout functions in Tensorflow are used for learning CNN models. The computer used for the experiments is equipped with 3.6 GHz quad-core processors and NVIDIA GTX980 GPU. 

A total of 404,200 training reps from 44,240 sets are trained with 100-sized minibatches. Also, 45,060 reps from 4,954 unseen sets serve as the test set. Note that reps from a single set belong to either the training or the test set, i.e., training and test sets are from separate sets. The data are standardized over the entire training dataset to have zero-mean and a variance of one in preprocessing. No additional preprocessing such as filtering or frame-aligning is applied to the raw data.  In particular, we do not perform any normalization to attempt to correct for differences in sensor placement or alignment between users.

\subsection{Effects of Image Formatting for Convolutions}

As explained in \ref{sec_CNNArchi}, there are three choices for shaping the input data: \texttt{84*84}, \texttt{9*784} and \texttt{3*784*3}. Note that depending on the image formatting, features which are separated in the input data vector may never have a chance to be convolved together until the last convolution layer. For example, in the \texttt{9*784 full} experiment, \texttt{Acc\_x\_local}, located in the top row of the image, will not be convolved with \texttt{EulerAngle\_z}, located in the bottom row, until the last layer. On the other hand, \texttt{Acc\_z\_world}, located in the 6th row in the image, will have many chances to be convolved with \texttt{EulerAngle\_x}, located in the 7th row, because they are adjacent to each other.

To remove this bias posed by the location of features in the input data vector, we may separate feature groups so that no convolution happens between the groups in the first layer. This is achieved by a CNN with \texttt{3*784*3} images. In \texttt{3*784*3} images, convolutions will happen only within a feature group in the first layer, and their contributions will be summed up to create new feature maps for the second layer. 

Another approach to avoid inter-group convolutions in the first layer is realized in the \texttt{9*789 disjointed} model. In this case, the image format is the same as \texttt{9*789 full}, however, convolution windows jump from a feature group to another feature group by taking a stride size of three (See Fig. \ref{fig:Overlap}). Therefore, the number of features will be reduced from 9 to 3 after the first layer and these 3 features will be convolved in the following layers.

The results after training for 40 epochs with these different image formats are presented in Table \ref{Table1}. For the data which have more than 7 reps, the best test result is 90.47\%, achieved by the \texttt{3*3*784} model while the worst test result is 88.85\%, achieved by the \texttt{9*784 full} model. The results show that convolutions on disjointed feature groups (\texttt{9*784 disj} and \texttt{3*784*3}) provide slightly better results than convolutions over full feature groups (\texttt{84*84} and \texttt{9*784 full}). In other words, convolutions between different feature groups (e.g. \texttt{Acc\_y, Acc\_z} and \texttt{EulerAngle\_x}) have little benefit for the classification task. These results are consistent with what we expected: physically meaningful convolutions can create better features than random convolutions.

Fig. \ref{fig:Confusion} presents the confusion matrix for the 3*784*3 experiments in rep-based predictions. The most easily classifiable exercises are \texttt{6:Wide-grip back lat pull-down with pulley machine} (Train: 99.86\%, Test: 98.08\%) and \texttt{0:Standing triceps extension with dumbbell} (99.90\%, 96.21\%) while the hardest are \texttt{43:Declined bench press with barbell} (88.06\%, 51.71\%) and \texttt{25:Alternating Romanian deadlift with dumbbell} (90.90\%, 56.77\%). In particular, the chance that the exercise 25 is misclassified as the exercise 43 is 41.69\%. One possible reason for this is that they have relatively weak signals and similar lengths. Indeed, the exercise 43 and 25 are the exercises which have the first and third smallest signal magnitudes among the 50 exercises and similar rep lengths which are $2.00$ and $2.11$ seconds on average.

\subsection{Effects of Data Reps and CNN Depth}

\begin{figure}[!t]
	\centering
    \includegraphics[trim={0cm 6cm 0cm, 0cm},clip,scale=0.25]{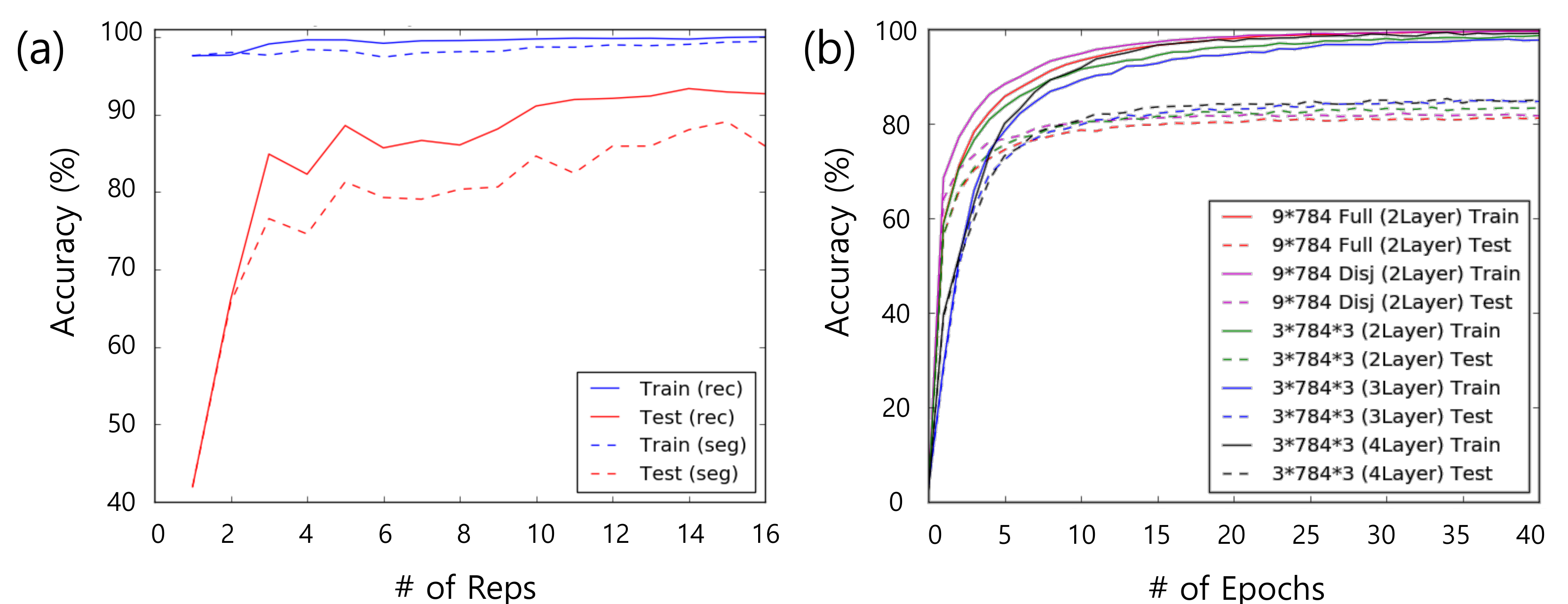}
    \caption{(a) Performance comparison between small-rep and many-rep exercises. (b) Learning curves with different CNN architectures. }
    \label{fig:Epoch}
\end{figure}

As presented in Table \ref{Table1}, the data which have many reps show better performance than the data which have few reps. This is to be expected in set-based predictions because if there are many reps in a set, then the final decision can be made based on the votes of many rep-based predictions. On the other hand, in extreme cases, single-rep sets cannot take the benefit of voting, thus, they obtain the same performance in rep-based and set-based predictions.

An interesting observation is that many-rep data show better performance than few-rep data even in rep-based predictions (Fig. \ref{fig:Epoch}(a)). The reps from 1-rep exercises have 42.32\% test accuracy while the reps from 20-rep exercises have 97.23\% accuracy. A possible explanation for this result is that people perform more consistent movements when they perform many-rep exercises with a light load, while they tend to demonstrate more explosive and fluctuating movements when they perform few-rep exercises with a heavier load. Indeed, the average loads for 1-rep and 20-rep exercises are 51.1kg and 20.9kg, while the variances of the all reps of 1-rep and 20-rep exercises are 5.59 and 0.94, respectively.

Fig. \ref{fig:Epoch}(a) shows that trained knowledge from few-rep exercises fails to be generalized to other unseen trials. This can be due to a large within-class variability in few-rep exercises, as we observed a large variance value from 1-rep exercise data. We may avoid this problem by reporting classification results only when many reps are observed. However, few-rep exercise data should not be ignored because many athletes often train with few-rep exercises for building their muscle strength. In the \emph{PUSH} dataset, 36\% of the sets have fewer than 8 reps and 9\% of the sets have fewer than 4 reps. Thus, improving the classification performance for few-rep exercises will be the next challenge in our exercise recognition problem.

To improve the generalization ability of the network, we attempted to increase the depth of the CNN. In this experiment, the 3-layer (92.14\%) and 4-layer (92.08\%) CNNs show better test results than the 2-layer (90.47\%) CNN (Table \ref{Table2})). From these results, it appears that additional layers provide some performance improvement. However, CNNs with more than 3 layers fail to outperform the 3-layer CNN. A direction for future work is to develop a customized deep CNN architecture which improves performance over the standard 3-layer CNN, particularly for small-rep exercises. The learning curves with different CNN architectures and image formats are presented in Fig. \ref{fig:Epoch}(b).

\begin{table}[!t]
\centering
\caption{Learning results with the \texttt{3*784*3} inputs  with different depths of CNNs. Deeper networks show better generalization ability, if large enough data are available.}
\label{Table2}
\begin{tabular}{@{}cccccc@{}}
\toprule
Layers & Feature maps    &   Dataset    & Rep   & Set   & \begin{tabular}[c]{@{}c@{}}Rec\\ (\textgreater7reps)\end{tabular} \\ \midrule
\multirow{2}{*}{2} & 32-64         & Train & 98.64 & 99.53 & 99.75  \\
    &   -(1024)-(1024)     & Test  & 83.34 & 87.04 & 90.47  \\
\multirow{2}{*}{3} & 32-64-128     & Train & 98.49 & 99.39 & 99.78  \\
    &   -(1024)-(1024)     & Test  & \textbf{85.26} & \textbf{88.55} & \textbf{92.14}  \\
\multirow{2}{*}{4} & 32-64-128-256 & Train & 99.12  &  99.65  & 99.86   \\
    &  -(2048)-(1024)      & Test  &  84.98 & 88.41 & 92.08   \\ \midrule
\end{tabular}
\end{table}
 
\section{Conclusion} \label{sec_conclusion}

In this paper, we propose an approach for classifying large-scale wearable sensor data of exercise movements using CNN and demonstrate 92.14\% classification accuracy on a 50-class exercise dataset with the 3-layer CNN. Experimental results empirically indicate that treating different feature groups - local acceleration, world acceleration, Euler angle in this case - as different channels of images \texttt{(3*784*3)} gives better results than providing 2D square \texttt{(84*84)} images or rectangle \texttt{(9*783)} images. Also, deeper CNNs give better results than shallow CNNs, although further research is required to fully exploit the benefit of a deeper structure. Lastly, sets with a large number of reps are easier to classify than ones with a few reps because people tend to perform movements more consistently when they perform a large number of reps of an exercise.

The current research used pre-segmented data which have relatively smaller within-class variance than nonsegmented data. The pre-segmented data also ease the variable length problem of time-series data so that CNN can treat the data as fixed-size images using simple zero-padding. In future work, we will investigate classification without segmentation, by introducing deep CNN or combining with neural networks for time-series data, e.g., LSTM \cite{LSTM1997}.






\section*{APPENDIX}
\subsection{List of Exercises for PUSH dataset}
\noindent
\it{(Sorted by the number of reps)}\\
\small
0. Standing triceps extension with dumbbell\\
1. Alternating lunges with dumbbell\\
2. Hammer-curl with dumbbell\\
3. Underhand-grip bent-over row with barbell\\
4. Lying triceps extension with dumbbell\\
5. Rope triceps push-down with pulley machine\\
6. Wide-grip back lat pull-down with pulley machine\\
7. Alternating backward lunges with dumbbell\\
8. Inclined bench press with dumbbell\\
9. Preacher curl with EZcurl bar\\
10. Side-raise with dumbbell\\
11. Triceps push-down with pulley machine\\
12. Push-ups\\
13. Wide-grip front lat pull-down with pulley machine\\
14. Kettlebell swing\\
15. Front-raise with dumbbell\\
16. Right-handed triceps Kick-back with dumbbell\\
17. Bench fly with dumbbell\\
18. Alternating traveling lunges with dumbbell\\
19. Reverse fly with dumbbell\\
20. Narrow-grip lat pull-down with pulley machine\\
21. Bench press with dumbbell\\
22. Alternating lunges with barbell\\
23. Seated military press with dumbbell\\
24. Goblet squat with dumbbell\\
25. Alternating Romanian deadlift with dumbbell\\
26. Bicep curl with dumbbell\\
27. Bent-over row with barbell\\
28. Left-handed split squat with barbell\\
29. Right-handed one arm row with dumbbell\\
30. Curl with EZcurl bar\\
31. Bent-over row with dumbbell\\
32. Romanian deadlift with barbell\\
33. Upright row with barbell\\
34. Hip thrust with barbell\\
35. Standing military press with dumbbell\\
36. Inclined fly with dumbbell\\
37. Inclined bench press with barbell\\
38. Alternating grip bent-over row with barbell\\
39. Arnold press with dumbbell\\
40. Goblet squat with kettlebell\\
41. Overhead press with barbell\\
42. Barbell good-morning\\
43. Declined bench press with barbell\\
44. Alternating step-ups with dumbbell\\
45. Bicep curl with pulley machine\\
46. Narrow-grip bench press with barbell\\
47. Wide-grip inclined bench press with barbell\\
48. Right-handed split squat with barbell\\
49. Wide-grip bench press with barbell

\subsection{List of Exercises Used in RecoFit Research \cite{RecoFit}} 
\noindent
\small
0. Crunch\\
1. Row\\
2. Punch\\
3. Jumping jack\\
4. Kettlebell swing\\
5. Triceps extension\\
6. Push-up\\
7. Rowing machine\\
8. Russian twist\\
9. Back fly\\
10. Shoulder press\\
11. Squat\\
12. Curl\\

\section*{ACKNOWLEDGMENT}

This research was supported by Canada's Natural Sciences and Engineering Research Council. We thank Rami Alhamad and \emph{PUSH Inc.}, who provided the wearable sensor data for the research.

\bibliographystyle{IEEEtran}
\bibliography{IEEEabrv,ICRA2017_PUSH_Terry}

\end{document}